# Adversarial Attacks on Image Classification Models – FGSM and Patch Attacks and their Impact

Jaydip Sen and Subhasis Dasgupta
Department of Data Science
Praxis Business School, Kolkata, India
Corresponding author's email: jaydip.sen@acm.org.

**Abstract**

This chapter introduces the concept of adversarial attacks on image classification models built on convolutional neural networks (CNN). CNNs are very popular deep-learning models which are used in image classification tasks. However, very powerful and pre-trained CNN models working very accurately on image datasets for image classification tasks may perform disastrously when the networks are under adversarial attacks. In this work, two very well-known adversarial attacks are discussed and their impact on the performance of image classifiers is analyzed. These two adversarial attacks are the fast gradient sign method (FGSM) and adversarial patch attack. These attacks are launched on three powerful pre-trained image classifier architectures, ResNet-34, GoogleNet, and DenseNet-161. The classification accuracy of the models in the absence and presence of the two attacks are computed on images from the publicly accessible ImageNet dataset. The results are analyzed to evaluate the impact of the attacks on the image classification task.

## 1. Introduction

Szegedy et al. observed that a number of machine-learning models, even cutting-edge neural networks, are susceptible to adversarial samples [1]. In other words, these machine learning models categorize incorrectly cases that differ by a marginal amount from examples that are correctly classified and taken from the distribution of data. The same adversarial example is most often classified incorrectly by a wide range of models of varied architecture which are built on different sub-samples of the training data. This shows that fundamental flaws in our training algorithms are exposed by adversarial samples. It was unclear what caused these adversarial cases. However, speculative explanations have indicated that it may be related to the extreme non-linearity property of deep neural networks in combination with inadequate model averaging and insufficient regularisation of the supervised learning problem that the models attempt to handle.

However, Goodfellow et al. disprove the need for these speculative hypotheses [2]. The authors argued that only linear behavior in high-dimensional domains is needed to produce adversarial cases. With the help of this viewpoint, it is possible to quickly create adversarial examples, which makes adversarial training feasible. The authors have also demonstrated that adversarial training can also regularize deep learning models in addition to the regularisation benefits offered by the techniques such as dropout [3]. Changing to nonlinear model families like RBF networks can significantly reduce a model's vulnerability to adversarial examples compared to generic regularisation procedures like dropout, pretraining, and model averaging.

One may consider deep learning, which is frequently employed in autonomous (driverless) automobiles, to see why such misclassification is risky [4]. To recognize signs or other cars on the road, systems based on DNNs are utilized [5]. The automobile might not stop and end up in a collision, which might have disastrous repercussions, if tampering with the input of such systems, for as by significantly changing the body of the car, stops DNNs from correctly recognizing it as a moving vehicle. When an enemy may gain by avoiding detection or having their information misclassified, there is a significant threat. These kinds of attacks are frequent in non-DL classification systems nowadays [6-10].

Goodfellow et al. argue that there is a fundamental incompatibility between building simple-to-train linear models and building models that use nonlinear effects to withstand hostile disruption [2]. By creating more effective optimization methods that can successfully train more nonlinear models in the long run, this trade-off may be avoided.

While the bulk of adversarial attacks has concentrated on slightly altering each pixel of an image, there are examples of attacks that are not limited to barely discernible alterations in the image. An approach that is based on creating an image-independent patch and positioning it to cover a tiny area of the image was demonstrated by Brown et al. [11]. The classifier will reliably predict a particular class for the image in the presence of this patch based on the attacker's preference. This assault is significantly more dangerous than pixel-based attacks like FGSM because it can potentially cause even more damage and because the attacker does not need to know what image they are attacking when they are building the attack. An adversarial patch might then be produced and disseminated for use by more attackers. The conventional defense strategies, which concentrate on protecting against minor perturbations, may not be robust to larger disturbances like these since the attack involves a massive perturbation.

This chapter discusses various adversarial attacks on image classification models and focuses particularly on two specific attacks, the *fast gradient sign method* (FGSM), and the *adversarial patch attack*. The impact of these two attacks on image classification accuracy is analyzed and extensive results are presented. The rest of the chapter is organized as follows. Section 2 presents a few related works. Some theoretical background information on adversarial attacks and pre-trained image classification models is discussed in Section 3. Section 4 presents detailed results and their analysis. Finally, the chapter is concluded in Section 5 highlighting some future works.

## 2. Related Work

Deep learning systems are generally prone to adversarial instances. These instances are deliberately selected inputs that influence the network to alter its output without being obvious to a human [5, 17]. Several optimization techniques, including L-BFGS [1], Fast Gradient Sign Method (FGSM) [2], DeepFool [12], and Projected Gradient Descent (PGD) [13] can be used to find these adversarial examples, which typically change each pixel by only a small amount. Other attack strategies aim to change only a small portion of the image's pixels (Jacobian-based saliency map [14]), or a small patch at a predetermined location [15].

A wide range of fascinating traits of neural networks and related models were demonstrated by Szegedy et al. [1]. The following are some of the important observations of the study: (1) Box-constrained L-BFGS can consistently discover adversarial cases. (2) The adversarial instances in ImageNet [16] data are so similar to the original examples that it is impossible for a human to distinguish between the two. (3) The same adversarial example is commonly classified incorrectly by a large number of classification models, each of which is trained using a different sample of the training data. (4) Adversarial events sometimes make shallow Softmax regression models less robust. (5) Training on adversarial examples can lead to a better regularization of the classification models.

By printing out a huge poster that resembles a stop sign or by applying various stickers to a stop sign, Eykholt et al. [17] demonstrated numerous techniques for creating stop signs that are misclassified by models.

These results suggest that classifiers developed using modern machine learning methods do not actually learn the underlying principles that determine the appropriate output label, even if they perform exceptionally well on the test data. The classification algorithms working for these models perform flawlessly with naturally occurring data. However, their classification accuracy drastically reduces for points that have a low probability in the underlying data distribution. This poses a big challenge to image classification since convolutional neural networks used in the classification work on the computation of perceptual feature similarity based on Euclidean distance. However, the resemblance found in this approach is false if images with unrealistically small perceptual distances actually belong to different classes as per the representation of the neural network.

The problem discussed above is particularly relevant to deep neural networks although linear classifiers are not immune to this problem. No model has yet been able to resist adversarial perturbation while preserving state-of-the-art accuracy on clean inputs. However, several approaches to defending against small perturbations-based adversarial attacks and some novel training approaches have been proposed by researchers [13, 14, 18, 19-26]. Some of these works proposing methods to defend against adversarial attacks are briefly presented in the following.

Madry et al. designed and trained deep neural networks on the MNIST and CIFAR10 image set that are robust to a wide range of adversarial attacks [13]. The authors formulated an approach to identify a saddle point for optimizing the error function and used a projected gradient descent (PGD) as the adversary. The proposed approach was found to yield a classification accuracy of 89% against the strongest adversary in the test data.

Papernot et al. proposed a novel method to create adversarial samples based on a thorough comprehension of the mapping between inputs and outputs of deep neural networks [14]. In a computer vision application, the authors demonstrated that, while only changing an average of 4.02% of the input characteristics of each sample, their proposed method can consistently create samples that were correctly classified by humans but incorrectly classified in certain targets by a deep neural network with a 97% adversarial success rate. Then, by designing a hardness metric, the authors assessed the susceptibility of various sample classes to adversarial perturbations and outlined a defense mechanism against adversarial samples.

Tramer et al. observed that adversarial attacks are more impactful in a black-box setup, in which perturbations are computed and transferred on undefended models [18]. Adversarial attacks are also very effective when they are launched in a single step that escapes the non-smooth neighborhood of the input data through a short random step. The authors proposed an ensemble adversarial training, a method that adds perturbations obtained from other models to training data. The proposed approach is found to be resistant to black-box adversarial attacks on the ImageNet dataset.

For assessing adversarial resilience on image classification tasks, Dong et al. developed a reliable benchmark [21]. The authors made some important useful observations including the following. First, adversarial training is one of the most effective defense strategies because it can generalize across different threat models. Second. model robustness ness curves are useful in the evaluation of the adversarial robustness of models. Finally, the randomization-based defenses are more resistant to query-based black-box attacks.

Chen et al. examined and evaluated the features and effectiveness of several defense strategies against adversarial attacks [22]. The authors considered the evaluation from four different perspectives: (i) gradient masking, (ii) adversarial training, (iii) adversarial examples detection, and (iv) input modifications. The authors presented several benefits and drawbacks of various defense mechanisms against adversarial attacks and explored the future trends in designing robust methods to defend against such attacks on image classification models.

## 3. Background Concepts

In this section, for the benefit of the readers, some background theories are discussed. The concepts of adversarial attack, fast gradient sign method (FGSM) attack, and three pre-trained convolutional neural network (CNN)-based deep neural network models, ResNet-34, GoogleNet, and DenseNet-161, are briefly introduced in this section.

### 3.1 Adversarial Attacks

Many different adversarial attack plans have been put out, all of which aim to significantly affect the model's prediction by slightly changing the data or picture input. How can we modify the image of a goldfish so that a classification model that could correctly classify the image before would no longer recognize it? On the other hand, a human would still categorize the image as a goldfish without any doubt, hence the label of the image shouldn't change at the same time. The generator network's goal under the framework for generative adversarial networks is the same as this one: try to trick another network (a discriminator) by altering its input.

### 3.2 Fast Gradient Sign Method

The Fast Gradient Sign Method (FGSM), created by Ian Goodfellow et al., is one of the initial attack tactics suggested [2]. The FGSM uses a neural network's gradients to produce an adversarial image. Essentially, the adversarial image is produced by FGSM by computing the gradients of a loss function (such as mean-square error or category cross-entropy) with respect to the input image and using the sign of the gradients to produce a new image (i.e., the adversarial image) that maximizes the loss. The end result is an output image that, to human sight, appears just like the original, but it causes the neural network to anticipate something different than it should have. The FGSM is represented in (1).

$$adv_x = x + \varepsilon * sign(\nabla_x J(\theta, x, y)) \quad (1)$$

The symbols used in (1) have the following significance:

$adv_x$     the adversarial image as the output

$x$     the original image as the input

$y$     the actual class (i.e., the ground-truth label) of the input image

$\varepsilon$     the noise intensity expressed as a small fractional value by which the signed gradients are multiplied to create perturbations. The perturbations should be small enough so that the human eye cannot distinguish the adversarial image from the original image.

$\theta$     the neural network model used for image classification

$J$     the loss function

The FGSM attack on an image involves the following three steps.

1. The value of the loss function is computed after the forward propagation in the network.
2. The gradients are computed with respect to pixels in the original (i.e., input) image.
3. The pixels of the input image are perturbed slightly in the direction of the computed gradients so that the value of the loss function is maximized.

Most often, in machine learning, determining the loss after forward propagation is frequently the initial step. To determine how closely the model's prediction matches the actual class, a negative likelihood loss function is used. Gradients are used to choose the direction in which to move the weights in order to lower the value of the loss function when training neural networks. However, calculating gradients in relation to an image's pixels is not a usual task. In FGSM, the pixels in the input image are moved in the direction of the gradient to maximize the value of the loss function.

### 3.3 ResNet-34 Architecture

He et al. presented a cutting-edge image classification neural network model containing 34 layers [27]. This deep convolutional neural network is known as the ResNet-34 model. The ImageNet dataset, which includes more than 100,000 images in 200 different classes, served as the pre-training data for ResNet-34. Similar to residual neural networks used for text prediction, ResNet architecture differs from typical neural networks in that it uses the residuals from each layer in the connected layers that follow.

### 3.4 GoogleNet Architecture

Szegedy et al. introduced GoogleNet (also known as Inception V1) in their paper titled "Going Deeper with Convolutions" [28]. In the 2014 ILSVRC image classification competition, this architecture was the winner. This architecture employs methods like global average pooling and 1–1 convolution in the middle of the architecture. A network may experience the issue of overfitting if it is constructed with very deep layers. To address this issue, the GoogleNet architecture was developed with the idea of having filters of various sizes that could function at the same level [28]. The network actually gets bigger with this concept rather than deeper. The architecture has a total of 22 layers, including 27 pooling layers. There are nine linearly stacked inception components that are connected to the global average pooling layer. The readers may refer to the work of Szegedy et al. for more details [28].

### 3.5 DenseNet-161 Architecture

A class of CNN called DenseNets uses dense connections between network layers for matching convolution operation feature-map sizes [29]. These dense connections are called dense blocks. Each layer receives extra inputs from all earlier layers and transmits its own feature maps to all later layers in order to maintain the feed-forward character of the system. Huang et al. demonstrated that a variant of DenseNet architecture called DensseNet-161with $k$ = 48 features per layer and having 29 million parameters can achieve a classification accuracy of 77.8% (i.e., top-1 classification accuracy) on the ImageNet ILSVRC classification dataset. As its name implies, the DenseNet-161 architecture contains 161 layers of nodes. More details on DenseNet-161 architecture may be found in [29, 30].

## 4. Image Classification Results and Analysis

Experiments are conducted to analyze the effect of two types of adversarial attacks on three well-known pre-trained CNN architectures. Two adversarial attacks considered in the study are the *FGSM attack* and the *adversarial patch attack* on a set of images. Three pre-trained architectures on which the attacks are simulated are ResNet-34, GoogleNet, and DenseNet-161. The images are chosen from the ImageNet dataset [16]. The pre-trained CNN models of ResNet-34, GoogleNet, and DenseNet-161 integrated into PyTorch's *torchvision* package, are used in the experiments.

### 4.1 Classification results in the absence of an attack

Before we study the impact of adversarial attacks on the image classification models, we analyze the classification accuracy of the models in the absence of any attack. Since the ImageNet dataset includes 1000 classes, it is not prudent to evaluate a model's performance just on the basis of its classification accuracy alone. Consider a model that consistently predicts the true label of an input image as the second-highest class using the *Softmax* activation function. Despite the fact that we would say it recognizes the object in the image, its accuracy is zero. There isn't always one distinct label we can assign an image to in ImageNet's 1000 classes. This is why "Top-5 accuracy" is a popular alternative metric for picture classification over a large number of classes. It shows how often the real label has been within the model's top 5 most likely predictions. Since the three pre-trained architectures perform very well on the images in the ImageNet dataset, instead of accuracy, the error, i.e., (1-accuracy) values are presented in the results.

Table 1 presents the performance results of three classification models on the whole ImageNet dataset containing 1000 classes of images. It is evident that all three models are highly accurate as depicted by their Top-% error percentage values. The DeepNet-161 model has yielded the highest level of accuracy and the least error among the three architectures. The Top-5 and Top-1 error rates for this model are found to be 2.30% and 15.10%, respectively.

**Table 1:** Classification accuracy of ResNet-34, GoogleNet, and Densenet-161 CNN models on the ImageNet data

| Metric | ResNet34 Model | GoogleNet Model | DenseNet161 Model |
|---|---|---|---|
| Top-1 error (%) | 19.10 | 25.26 | 15.10 |
| Top-5 error (%) | 4.30 | 7.74 | 2.30 |

**Table 2:** The classification results of the ResNet-34 model for the chosen images

| Image Index | Image True Class | Top-5 Predicted Classes and their Confidence | |
|---|---|---|---|
| | | Class | Confidence |
| 0 | **tench** | **tench** | **0.9817** |
| | | barracouta | 0.0095 |
| | | coho | 0.0085 |
| | | gar | 0.0002 |
| | | sturgeon | 0.0001 |
| 6 | **goldfish** | **goldfish** | **0.9982** |
| | | tench | 0.0005 |
| | | barracouta | 0.0005 |
| | | tailed frog | 0.0003 |
| | | puffer | 0.0002 |
| 13 | **great white shark** | **great white shark** | **0.9855** |
| | | tiger shark | 0.0109 |
| | | submarine | 0.0007 |
| | | sturgeon | 0.0006 |
| | | hammerhead | 0.0005 |
| 18 | **tiger shark** | **tiger shark** | **0.9118** |
| | | sturgeon | 0.0251 |
| | | great white shark | 0.0202 |
| | | puffer | 0.0192 |
| | | electric ray | 0.0038 |

After evaluating the overall performance of the three models, we investigate some specific images in the dataset. For this purpose, the images with the indices 0, 6, 13, and 18 are randomly chosen and how the model has classified these images are checked. The images corresponding to the four indices chosen belong to the classes "tench", "goldfish", "great white shark" and "tiger shark", respectively.

Table 2 presents the performance of the ResNet-34 model on the classification task for the four images. It is evident that the model has been very accurate in classification as the confidence associated with the true class of each of the four images is more than 90%. It may be noted that confidence here means the probability value that the model associates with the corresponding class. For example, the ResNet-34 model has yielded a confidence value of 0.9817 for the image whose true class is "tench" with the predicted class "tench", implying that the model has associated a probability of 0.9817 with its classification of the image to the class "tench".

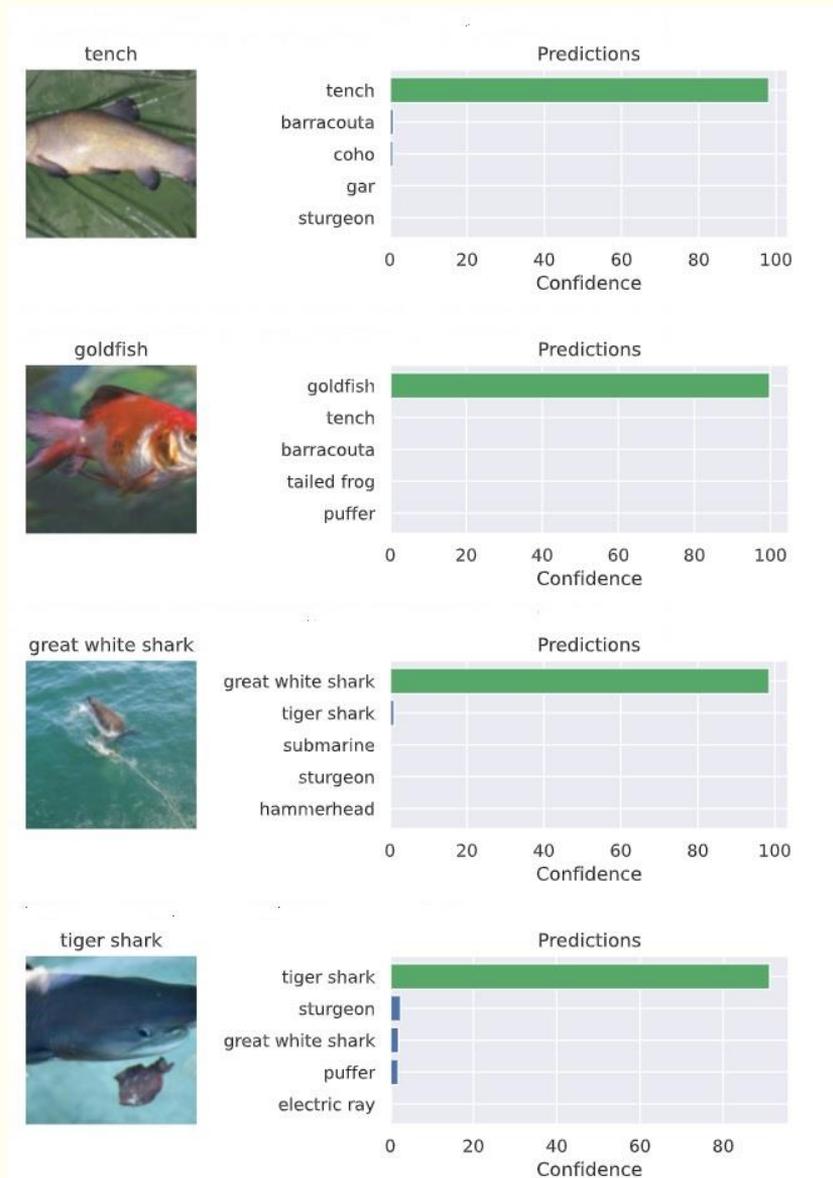

**Figure 1:** The classification results of the ResNet34 model for the chosen images

    **Figure 1** depicts the classification results of the ResNet-34 model on the four images. In Figure 1, the input image is shown on the left and the confidence values of the model for the top five classes for the image are shown on the right. The confidence values are shown in the form of horizontal bars.

**Table 3:** The classification results of the GoogleNet model for the chosen images

| Image Index | Image True Class | Top-5 Predicted Classes and their Confidence | |
|---|---|---|---|
| | | **Class** | **Confidence** |
| 0 | **tench** | **tench** | **0.9826** |
| | | coho | 0.0075 |
| | | barracouta | 0.0034 |
| | | goldfish | 0.0008 |
| | | gar | 0.0005 |
| 6 | **goldfish** | **goldfish** | **0.9617** |
| | | tench | 0.0129 |
| | | loggerhead | 0.0018 |
| | | barracouta | 0.0014 |
| | | coho | 0.0010 |
| 13 | **great white shark** | **great white shark** | **0.8188** |
| | | sea lion | 0.0532 |
| | | grey whale | 0.0376 |
| | | tiger shark | 0.0144 |
| | | loggerhead | 0.0082 |
| 18 | **tiger shark** | **tiger shark** | **0.3484** |
| | | platypus | 0.1978 |
| | | hammerhead | 0.0277 |
| | | sturgeon | 0.0245 |
| | | great white shark | 0.0166 |

**Table 3** presents the performance of the GoogleNet model on the classification task for the four images. It is observed that the model has been very accurate in the classification task for the "tench" and "goldfish" images. While its accuracy for the image "great white shark" class is high, the model has performed poorly for the image "tiger shark". However, for the "tiger shark" image the model has still associated the highest confidence value for the correct class, although the confidence is quite low, i.e., 0.3484.

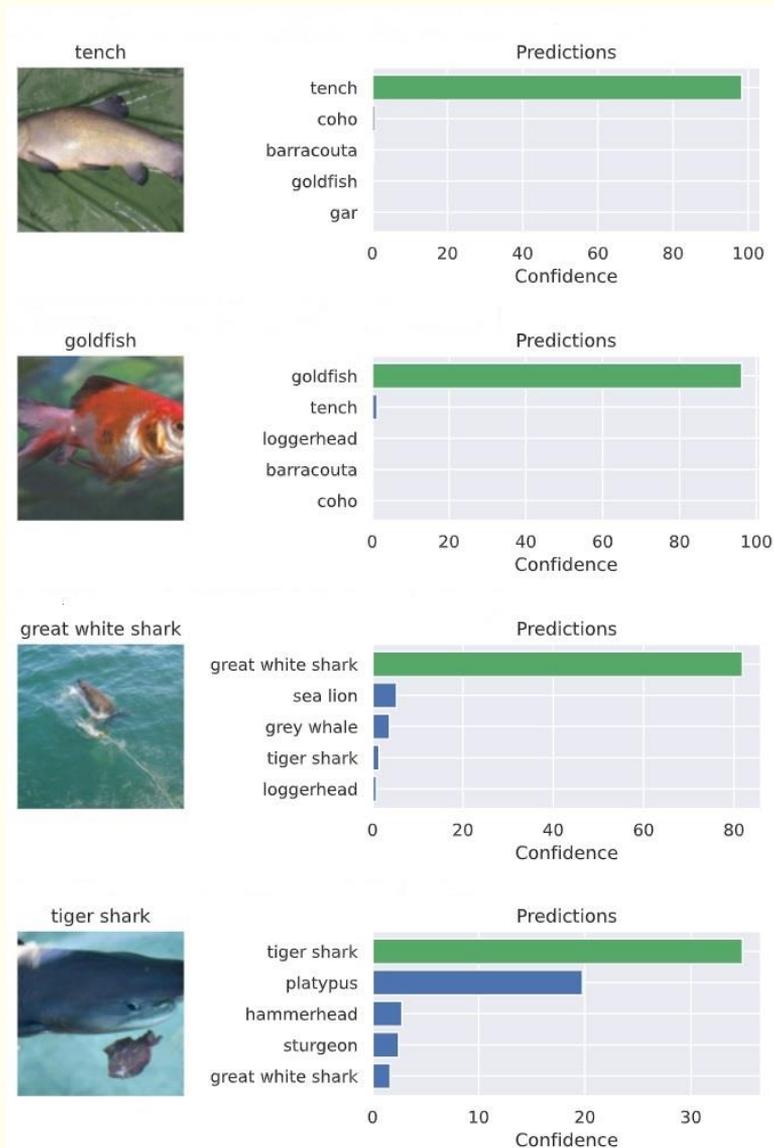

**Figure 2:** The classification results of the GoogleNet model for the chosen images

Figure 2 depicts the classification results of the GoogleNet model on the four images. In Figure 2, the input image is shown on the left and the confidence values of the model for the top five classes for the image are shown on the right. The confidence values are shown in the form of horizontal bars.

**Table 4:** The classification results of the DenseNet-161 model for the chosen images

| Image Index | Image True Class | Top-5 Predicted Classes and their Confidence ||
|---|---|---|---|
| | | **Class** | **Confidence** |
| 0 | **tench** | **tench** | **0.9993** |
| | | barracouta | 0.0003 |
| | | coho | 0.0002 |
| | | gar | 0.0001 |
| | | platypus | 0.0001 |
| 6 | **goldfish** | **goldfish** | **0.9999** |
| | | barracouta | 0.0001 |
| | | tench | 0.0001 |
| | | coho | 0.0001 |
| | | gar | 0.0001 |
| 13 | **great white shark** | **great white shark** | **0.9490** |
| | | tiger shark | 0.0177 |
| | | dugong | 0.0127 |
| | | sea lion | 0.0113 |
| | | grey whale | 0.0074 |
| 18 | **tiger shark** | **tiger shark** | **0.9932** |
| | | great white shark | 0.0047 |
| | | gar | 0.0008 |
| | | sturgeon | 0.0002 |
| | | hammerhead | 0.0001 |

**Table 4** presents the performance of the DenseNet-161 model on the classification task for the four images. It is observed that the performance of the model on the classification task has been excellent. For all four images, the confidence values computed by the model for the true class have been higher than 94. The results also show that among the three architectures, DenseNet-161 has been the most accurate model for the classification of the four images chosen for analysis.

**Figure 3** depicts the classification results of the DenseNet-161 model on the four images. In Figure 3, the input image is shown on the left and the confidence values of the model for the top five classes for the image are shown on the right. The confidence values are shown in the form of horizontal bars.

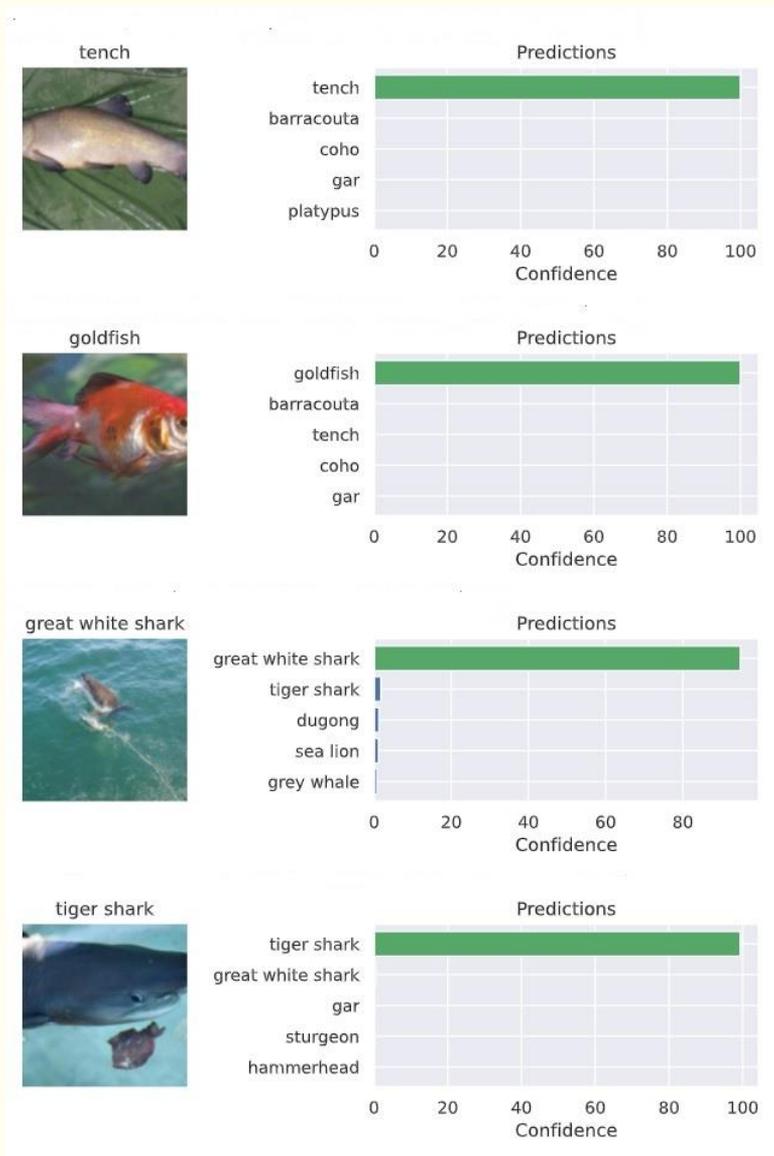

**Figure 3:** The classification results of the DenseNet161 model for the chosen images

## 4.2 Classification results in the presence of the FGSM attack

After observing the performance of the three CNN architectures for the image classification tasks on the images in the ImageNet dataset, the impact of the adversarial attacks on the classifier models is studied. We start with the FGSM attack with a value of 0.02 for epsilon ($\varepsilon$). The value of $\varepsilon = 0.02$ indicates that the values of pixels are changed by an amount of 1 (approximately) in the range of 0 to 255 – the range over which a pixel value can change. This change is so small that it will be impossible to distinguish the adversarial image from the original one. The performance results of the three models in the presence of FGSM attack with $\varepsilon = 0.02$ have been presented in Table 5 – Table 7. The results are pictorially depicted in Figure 4 – Figure 6.

**Table 5:** The performance of ResNet-34 model under FGSM attack with ε = 0.02

| Image True Class | Top-5 Predicted Classes and their confidence ||
|---|---|---|
| | **Class** | **Confidence** |
| **tench** | coho | 0.6684 |
| | barracouta | 0.2799 |
| | gar | 0.0321 |
| | **tench** | **0.0128** |
| | sturgeon | 0.0066 |
| **goldfish** | barracouta | 0.5885 |
| | tench | 0.1392 |
| | gar | 0.0945 |
| | tailed frog | 0.0462 |
| | coho | 0.0434 |
| **great white shark** | dugong | 0.2661 |
| | tiger shark | 0.2575 |
| | grey whale | 0.0578 |
| | **great white shark** | **0.0537** |
| | submarine | 0.0303 |
| **tiger shark** | otter | 0.2462 |
| | puffer | 0.1666 |
| | beaver | 0.1543 |
| | platypus | 0.1083 |
| | sea lion | 0.0565 |

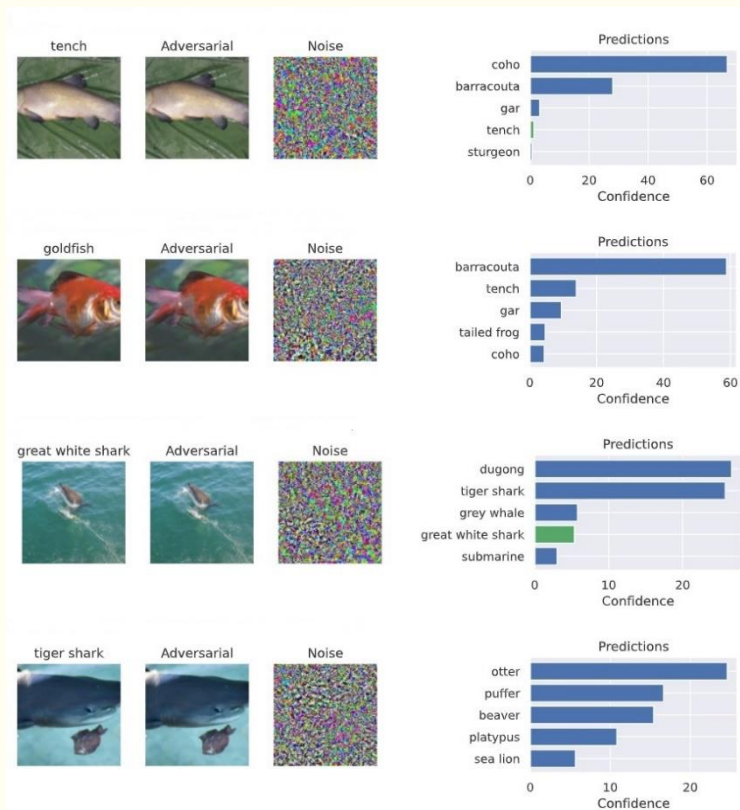

**Figure 4:** The performance of the ResNet-34 model under FGSM attack with ε = 0.02

**Table 6:** The performance of GoogleNet model under FGSM attack with ε = 0.02

| Image True Class | Top-5 Predicted Classes and their confidence | |
|---|---|---|
| | **Class** | **Confidence** |
| **tench** | coho | 0.2652 |
| | **tench** | **0.2116** |
| | barracouta | 0.1275 |
| | gar | 0.0219 |
| | sturgeon | 0.0153 |
| **goldfish** | **goldfish** | **0.0553** |
| | tench | 0.0305 |
| | barracouta | 0.0218 |
| | gar | 0.0127 |
| | great white shark | 0.0115 |
| **great white shark** | weasel | 0.1880 |
| | sea lion | 0.1489 |
| | otter | 0.1402 |
| | platypus | 0.0605 |
| | tailed frog | 0.0400 |
| **tiger shark** | platypus | 0.5450 |
| | beaver | 0.0336 |
| | American coot | 0.0265 |
| | terrapin | 0.0142 |
| | otter | 0.0127 |

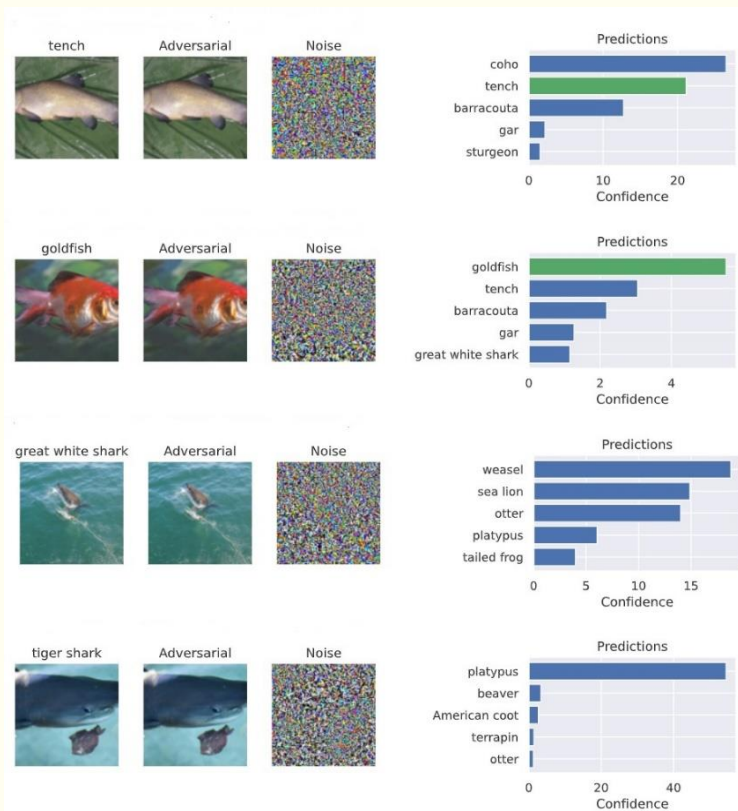

**Figure 5:** The performance of the GoogleNet model under FGSM attack with ε = 0.02

**Table 7:** The performance of DenseNet-161 model under FGSM attack with ε = 0.02

| Image True Class | Top-5 Predicted Classes and their confidence | |
|---|---|---|
| | **Class** | **Confidence** |
| **tench** | coho | 0.6793 |
| | **tench** | **0.1645** |
| | gar | 0.0466 |
| | barracouta | 0.0373 |
| | sturgeon | 0.0273 |
| **goldfish** | barracouta | 0.7139 |
| | tench | 0.1357 |
| | coho | 0.0665 |
| | gar | 0.0645 |
| | **goldfish** | **0.0137** |
| **great white shark** | sea lion | 0.4830 |
| | dugong | 0.4388 |
| | grey whale | 0.0255 |
| | tiger shark | 0.0099 |
| | snorkel | 0.0023 |
| **tiger shark** | great white shark | 0.9025 |
| | gar | 0.0606 |
| | barracouta | 0.0059 |
| | **tiger shark** | **0.0058** |
| | coho | 0.0058 |

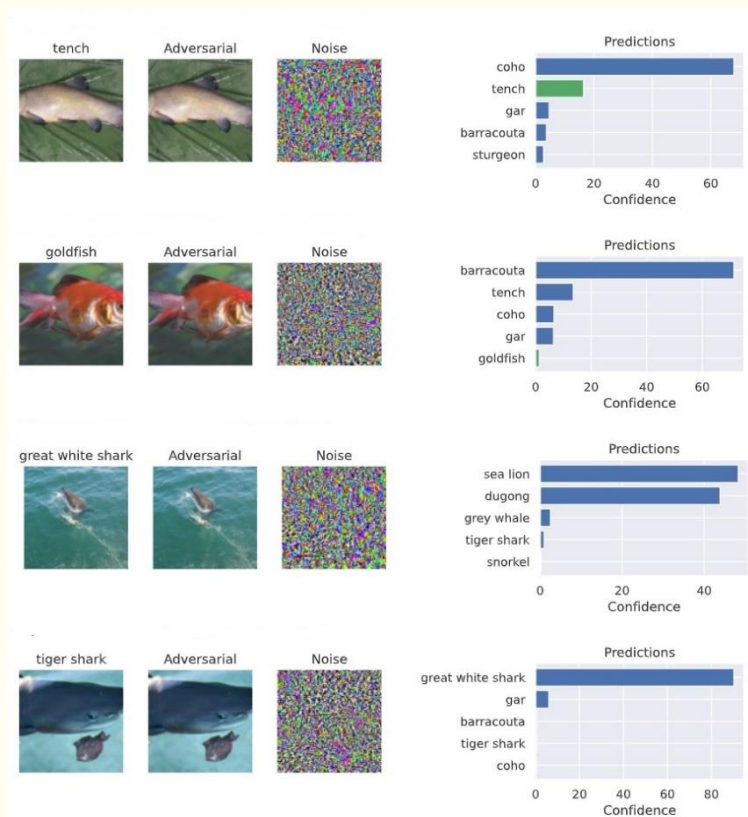

**Figure 6:** The performance of DenseNet-161 model under FGSM attack with ε = 0.02

It is evident that all three models are adversely affected by the FGSM attack even with a value of ε as low as 0.02. While the adversarial images are impossible to distinguish from the original ones, none of the models could correctly classify any of the four images as the highest confidence values were assigned to incorrect classes.

**Table 8:** Performance of ResNet34 under FGSM attack for different values of ε

| Noise Level (ε) | Classification Error in Percent | |
|---|---|---|
| | Top-1 Error | Top-5 Error |
| 0.01 | 83.44 | 43.76 |
| 0.02 | 93.56 | 60.54 |
| 0.03 | 95.66 | 68.60 |
| 0.04 | 96.24 | 72,42 |
| 0.05 | 96.76 | 74.78 |
| **0.06** | **97.00** | 76.18 |
| 0.07 | 96.98 | 76.92 |
| 0.08 | 97.00 | 77.54 |
| **0.09** | 96.94 | **77.68** |
| 0.10 | 96.92 | 77.56 |

**Table 9:** Performance of GoogleNet under FGSM attack for different values of ε

| Noise Level (ε) | Classification Error in Percent | |
|---|---|---|
| | Top-1 Error | Top-5 Error |
| 0.01 | 82.76 | 49.52 |
| 0.02 | 91.10 | 65.86 |
| 0.03 | 93.72 | 72.68 |
| 0.04 | 94.66 | 75.86 |
| 0.05 | 95.14 | 77.62 |
| 0.06 | 95.26 | 78.40 |
| 0.07 | 95.36 | 78.96 |
| 0.08 | 95.40 | 79.04 |
| **0.09** | **95.46** | **79.24** |
| 0.10 | 95.40 | 79.20 |

**Table 10:** Performance of DenseNet161 under FGSM attack for different values of ε

| Noise Level (ε) | Classification Error in Percent | |
|---|---|---|
| | Top-1 Error | Top-5 Error |
| 0.01 | 79.08 | 33.10 |
| 0.02 | 90.08 | 50.64 |
| 0.03 | 92.98 | 58.64 |
| 0.04 | 94.04 | 62.88 |
| 0.05 | 94.38 | 65.12 |
| 0.06 | 94.38 | 66.38 |
| 0.07 | 94.34 | 66.76 |
| **0.08** | **94.42** | **66.94** |
| 0.09 | 94.18 | 66.68 |
| 0.10 | 94.10 | 66.70 |

The value of the parameter ε is increased from 0.01 to 0.10 by a step of 0.01. It is observed that except for a few cases, the classification error increased consistently with ε till ε reaches a value in the range of 0.08 – 0.09. The impact of the FGSM attack is so severe that the classification error for the ResNet-34 model in the presence of this attack reaches as high values as 97.00% (Top-1 error) and 77.68% (Top-5 error). The corresponding values for the GoogleNet model are 95.46% (Top-1 error) and 79.24% (Top-5 error), and for the DenseNet-161 are 94.42% (Top-1 error) and 66.94% (Top-5 error). Among the three models, DenseNet-161 looked to be the most robust against the FGSM attack.

### 4.3 Classification results in the presence of the adversarial patch attack

As mentioned in Section 1, an attack can also be launched on image classification models by introducing adversarial patches [11]. In this attack, the strategy is to transform a small portion of the image into a desired form and shape instead of the FGSM's approach of slightly altering some pixels. This will be able to deceive the classification model and force it to predict a certain pre-determined class. In practical applications, this type of attack poses a greater hazard than FGSM. Consider an autonomous vehicle network that receives a real-time image from a camera. To trick this vehicle into thinking that an automobile is actually a pedestrian, another driver may print out a certain design and stick it on the rear part of the vehicle.

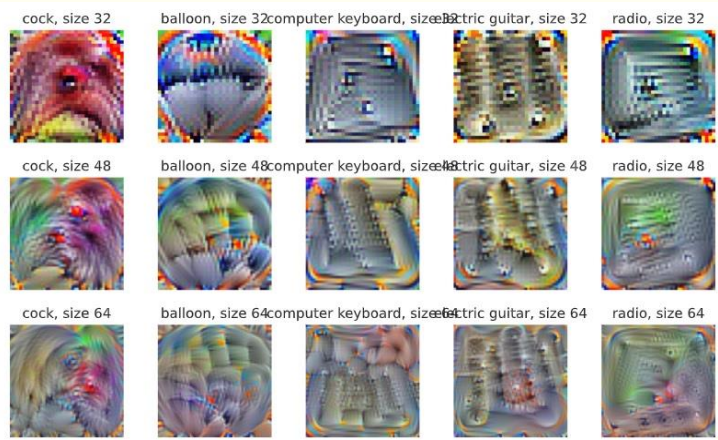

**Figure 6:** Five images used as patches: cock, balloon, computer keyboard, electric guitar, and radio. The sizes for the patch images: 32*32, 48*48, and 64*64.

For simulating the adversarial patch attack on the same four images on which the FGSM attack was launched, at first, five images are chosen randomly which will be used as the patches. The five patch images are (i) cock, (ii) balloon, (iii) computer keyboard, (iv) electric guitar, and (v) radio. For the purpose of studying the effect of the sizes of the patch images on the accuracies of the classification models, three different sizes are considered for each patch image. The three sizes are (i) 32*32, (ii) 48*48, and (iii) 64*64. The sizes are expressed in terms of the number of pixels along the x and y dimensions. Tables 11-16 present the accuracies (Top 1% and Top 5%) of the models for different sizes of different patch images. Here, accuracy refers to the percentage of cases in which the images have been classified as the target class (i.e., patch class) with the highest confidence. Figures 7 – 9 depict the performance of the classification models in the presence of a "balloon" patch of size 64*64. The pictures for other patch images and other sizes are not presented for the sake of brevity.

**Table 11:** Top-1 accuracy (%) of ResNet34 model for different patch sizes

| Patch Image | Size of the Patch Image | | |
|---|---|---|---|
| | 32*32 | 48*48 | 64*64 |
| cock | 78.75 | 92.01 | **97.82** |
| balloon | **81.17** | **92.35** | 97.44 |
| computer keyboard | 0.04 | 68.46 | 92.97 |
| electric guitar | 54.47 | 87.08 | 95.93 |
| radio | 22.35 | 75.49 | 94.03 |

**Table 12:** Top-5 accuracy (%) of ResNet34 model for different patch sizes

| Patch Image | Size of the Patch Image | | |
|---|---|---|---|
| | 32*32 | 48*48 | 64*64 |
| cock | 93.48 | 98.59 | **99.84** |
| balloon | **93.73** | **98.88** | 99.83 |
| computer keyboard | 1.22 | 91.63 | 99.45 |
| electric guitar | 77.43 | 97.15 | 99.64 |
| radio | 62.08 | 93.67 | 99.37 |

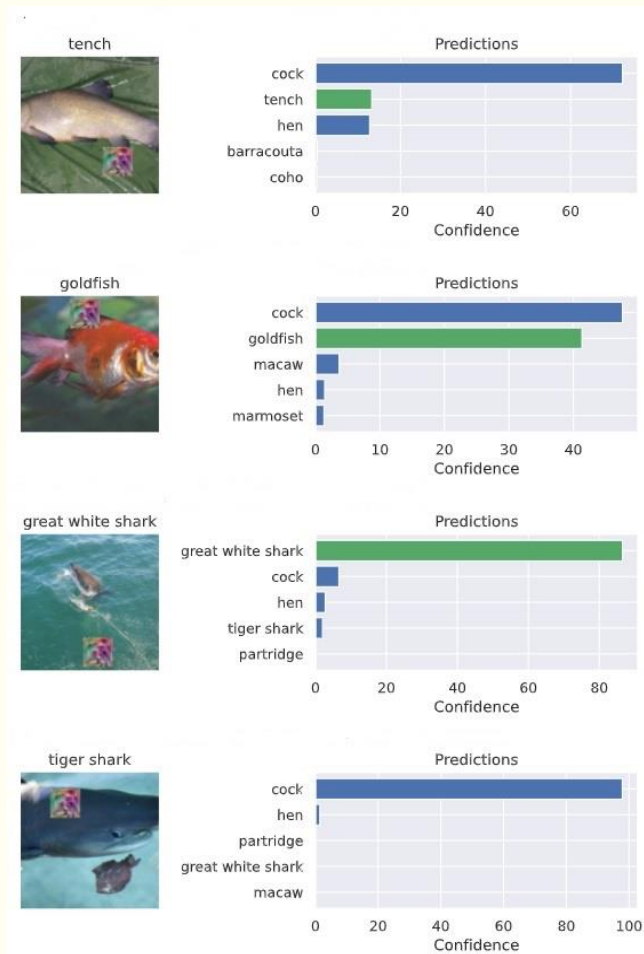

**Figure 7:** The classification results of the ResNet-34 model in the presence of a patch image of a balloon with size 64*64.

**Table 13:** Top-1 accuracy (%) of GoogleNet model for different patch sizes

| Patch Image | Size of the Patch Image | | |
|---|---|---|---|
| | 32*32 | 48*48 | 64*64 |
| cock | 0.00 | 0.27 | 0.94 |
| balloon | 85.06 | 95.67 | 98.76 |
| computer keyboard | 17.17 | 80.81 | 97.22 |
| electric guitar | 70.14 | 93.64 | 98.76 |
| radio | 7.73 | 81.34 | 95.59 |

**Table 14:** Top-5 accuracy (%) of GoogleNet model for different patch sizes

| Patch Image | Size of the Patch Image | | |
|---|---|---|---|
| | 32*32 | 48*48 | 64*64 |
| cock | 0.09 | 8.69 | 32.63 |
| balloon | 96.76 | 99.80 | 99.99 |
| computer keyboard | 66.72 | 96.81 | 99.94 |
| electric guitar | 90.45 | 99.60 | 99.98 |
| radio | 65.15 | 97.21 | 99.84 |

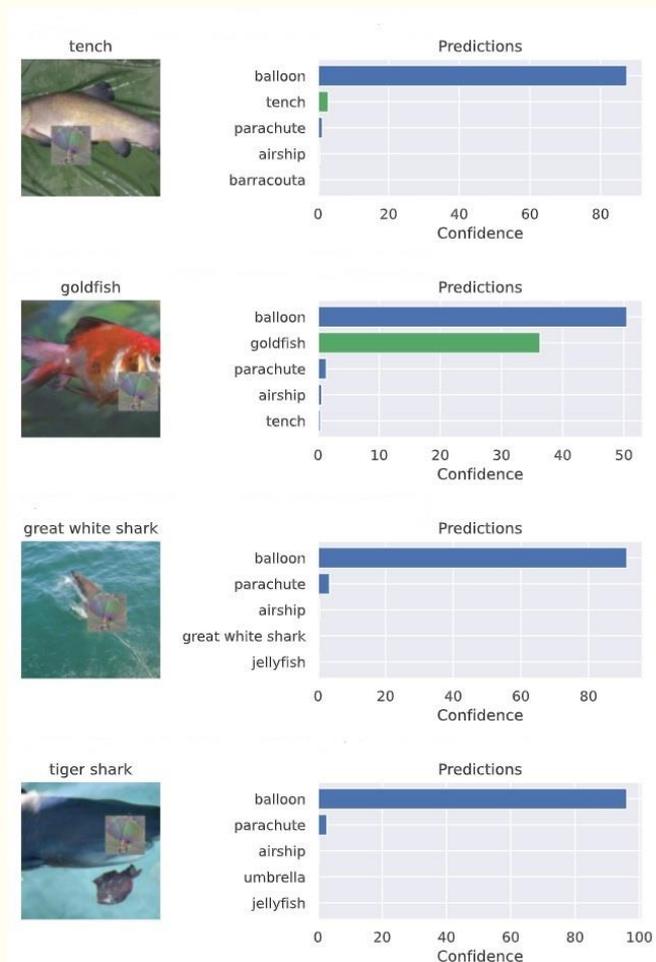

**Figure 8:** The classification results of the GoogleNet model in the presence of a patch image of a balloon with size 64*64.

**Table 15:** Top-1 accuracy (%) of DenseNet-161 model for different patch sizes

| Patch Image | Size of the Patch Image | | |
|---|---|---|---|
| | 32*32 | 48*48 | 64*64 |
| cock | 0.00 | 0.00 | 0.01 |
| balloon | **14.70** | **35.41** | 40.25 |
| computer keyboard | 0.02 | 0.08 | **47.46** |
| electric guitar | 1.91 | 5.97 | 46.74 |
| radio | 0.53 | 8.88 | 43.44 |

**Table 16:** Top-5 accuracy (%) of DenseNet161 model for different patch sizes

| Patch Image | Size of the Patch Image | | |
|---|---|---|---|
| | 32*32 | 48*48 | 64*64 |
| cock | 0.09 | 0.08 | 0.16 |
| balloon | **41.63** | **69.44** | 69.75 |
| computer keyboard | 0.76 | 1.46 | **79.86** |
| electric guitar | 13.10 | 22.67 | 75.43 |
| radio | 10.36 | 46.63 | 74.84 |

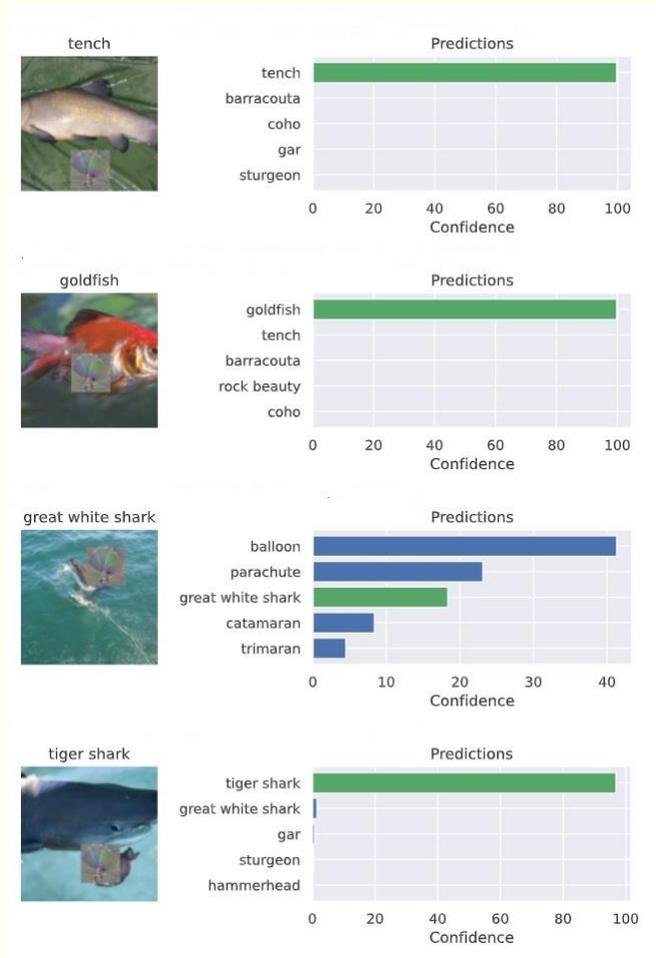

**Figure 9:** The classification results of the DenseNet-161 model in the presence of a patch image of a balloon with size 64*64.

The following observations are made on the results of the adversarial patch attack.

1. For the same patch image, all three models, ResNet-34, GoogleNet, and DenseNet-161, exhibited higher accuracy in deceiving the models into the wrong classification for a bigger patch size. In other words, for all three models, the attack effectiveness is the highest for the patch size 64*64, for a given patch image.
2. For most of the patch images and patch sizes, the effectiveness of the attack on three models is found to be the most for the patch image of "balloon". However, for the ResNet-34 model, with a patch image of "cock" the attack yielded the maximum effectiveness for the patch size of 64*64. For the GoogleNet model, along with the "balloon" patch image, the "electric guitar" patch of size 64*64 also produced the maximum Top-1 accuracy. For the DenseNet-161 model, the attack exhibited the highest effectiveness for the patch image of "computer keyboard" of size 64*64 for both Top-1 and Top-5 cases.
3. For obvious reasons, the attack effectiveness (i.e., the accuracy of the attack) is found to be always higher for the Top-5 case than its corresponding Top-1 counterpart.

## 7. Conclusion

In this chapter, some adversarial attacks on CNN-based image classification models are discussed. In particular, two attacks, e.g., the FGSM attack and adversarial patch attack are presented in detail. The former attack involves changing the pixels of an image in the direction of their maximum gradients so that the value of the loss function is maximized. While the resultant adversarial image is impossible to distinguish from the original image by human eyes, the highly trained classification models will most likely classify the adversarial image into a class that is different from its ground truth. For the adversarial patch attack, an image patch of a different class is inserted in the original image in such a way that the trained models will be deceived and forced to incorrectly classify the original image into the class of the patch image. It is observed in the study that with the increase in the amount of perturbation created in the original image by the FGSM attack, the error in the classification increases till a threshold level is reached at which the attack saturates. No further increase in perturbation usually leads to a further decrease in the classification accuracy of the models. For the adversarial patch attack, the attack effectiveness increases with the increase in the patch size.